\documentclass[letterpaper, 10 pt, conference]{ieeeconf}  % Comment this line out if you need a4paper

\IEEEoverridecommandlockouts                              % This command is only needed if 
\usepackage{graphics} % for pdf, bitmapped graphics files
\graphicspath{{./pdf/}} % load figure within subfolder ./pdf/
\DeclareGraphicsExtensions{.pdf} % for omit extension
\usepackage[ruled,linesnumbered]{algorithm2e}
\usepackage{epsfig} % for postscript graphics files
\usepackage{mathptmx} % assumes new font selection scheme installed
\usepackage{times} % assumes new font selection scheme installed
\usepackage{amsmath} % assumes amsmath package installed
\usepackage{amssymb}  % assumes amsmath package installed
\usepackage[table]{xcolor}

\usepackage{draftwatermark}% to add watermark
\usepackage{fancyhdr}% to add header
\usepackage{tikz}
\usepackage{textcomp}
\usepackage{hyperref}
\usepackage{lipsum}

\newcommand\copyrighttext{%
  \footnotesize \textcopyright 2012 IEEE. Personal use of this material is permitted.
  Permission from IEEE must be obtained for all other uses, in any current or future
  media, including reprinting/republishing this material for advertising or promotional
  purposes, creating new collective works, for resale or redistribution to servers or
  lists, or reuse of any copyrighted component of this work in other works.}
\newcommand\copyrightnotice{%
\begin{tikzpicture}[remember picture,overlay]
\node[anchor=south,yshift=10pt] at (current page.south) {\fbox{\parbox{\dimexpr\textwidth-\fboxsep-\fboxrule\relax}{\copyrighttext}}};
\end{tikzpicture}%
}

\newtheorem{theorem}{Theorem}
\title{\LARGE \bf
A Convex Optimization Framework for Constrained Concurrent Motion Control of a Hybrid Redundant Surgical System*
}

\author{Farshid Alambeigi$^{1}$, Shahriar Sefati$^{1}$, and Mehran Armand$^{1,2}$% <-this % stops a space
\thanks{*Research supported by NIH/NIBIB grant R01EB016703.}% <-this % stops a space
	\thanks{$^{1}$Farshid Alambeigi, Shahriar Sefati, and Mehran Armand are with Laboratory for Computational Sensing and Robotics, Johns Hopkins University, Baltimre, MD, USA, 21218. Email: \{falambe1,sefati, marmand2\}@jhu.edu}%
\thanks{$^{2}$Mehran Armand is also with Johns Hopkins University, Applied Physics Laboratory, Laurel, MD, USA, 20723. Email: mehran.armand@jhuapl.edu.}%
}

\begin{document}

%%%%%%%%%%%%%% TO ADD WAtermark
\SetWatermarkText{Accepted for ACC 2018}
\SetWatermarkScale{0.4}

\maketitle
\copyrightnotice
\thispagestyle{empty}
% * <alambeigi.farshid@gmail.com> 2017-09-24T16:51:44.793Z:
%
% ^.
\pagestyle{empty}

%%%%%%%%%%%%%%%%%%%%%%%%%%%%%%%%%%%%%%%%%%%%%%%%%%%%%%%%%%%%%%%%%%%%%%%%%%%%%%%%
\begin{abstract}

We present a constrained motion control framework for a  redundant surgical system designed for minimally invasive treatment of pelvic osteolysis. The framework comprises a kinematics model of a six Degrees-of-Freedom (DoF) robotic arm integrated with a one DoF continuum manipulator as well as a novel convex optimization redundancy resolution controller. 
To resolve the redundancy resolution problem, formulated as a constrained $\ell_2$-regularized quadratic minimization, we study and evaluate the potential use of an optimally tuned alternating direction method of multipliers (ADMM) algorithm. To this end, we prove global convergence of the algorithm at linear rate and propose expressions for the involved parameters resulting in a fast convergence. Simulations on the robotic system verified our analytical derivations and showed the capability and robustness of the ADMM algorithm in constrained motion control of our  redundant surgical system. 

\end{abstract}

%%%%%%%%%%%%%%%%%%%%%%%%%%%%%%%%%%%%%%%%%%%%%%%%%%%%%%%%%%%%%%%%%%%%%%%%%%%%%%%%
\section{INTRODUCTION}

Continuum Dexterous Manipulators (CDMs) have changed the paradigm of Minimally Invasive Surgical (MIS) procedures due to their added dexterity and compliancy compared to the common rigid surgical instruments \cite{simaan2009design}. These robots have been studied in a variety of surgical procedures interacting with soft tissues e.g. Natural OrificeTrans-luminal Endoscopic Surgery \cite{simaan2009design},\cite{kapoor2005suturing}, endoscopic submucosal dissection \cite{patel2015evaluation}, and petrous apex lesions in the skull base \cite{coemert2016integration}. Our group is currently focusing on developing a surgical system for MIS treatment of hard tissues and particularly treatment of osteolytic lesions behind the hip acetabular implant using a custom designed CDM \cite{alambeigi2016design},\cite{alambeigi2016toward} and Fiber Bragg Grating (FBG) optical sensors \cite{sefati2016fbg}, \cite{8234018}.

As shown in Fig. 1, in our proposed approach for MIS removing and treatment of osteolytic lesions behind the confined area of the implant, the CDM is first inserted through the screw holes of the well-fixed acetabular implant (with 8 mm diameter), and then controlled to the desired locations behind the implant. To this end, we utilize a hybrid redundant surgical system including a six Degrees-of-Freedom (DoF) robotic manipulator, as the positioning robot, and a one DoF CDM, as the dexterous guiding channel for different instruments (e.g. curettes, burs, and shaving tools). Of note, the used CDM is made of nitinol tubes (with outer diameter of 6 mm and inner diameter of 4 mm) with notches along its length, which constrains its bending to a single plane via two actuating tendons \cite{alambeigi2014control}. Successful motion control of this redundant surgical system in real time considering the safety concerns demands a fast enough and reliable concurrent control framework.
\begin{figure}[!t]
	\centering
	\includegraphics[width=\linewidth]{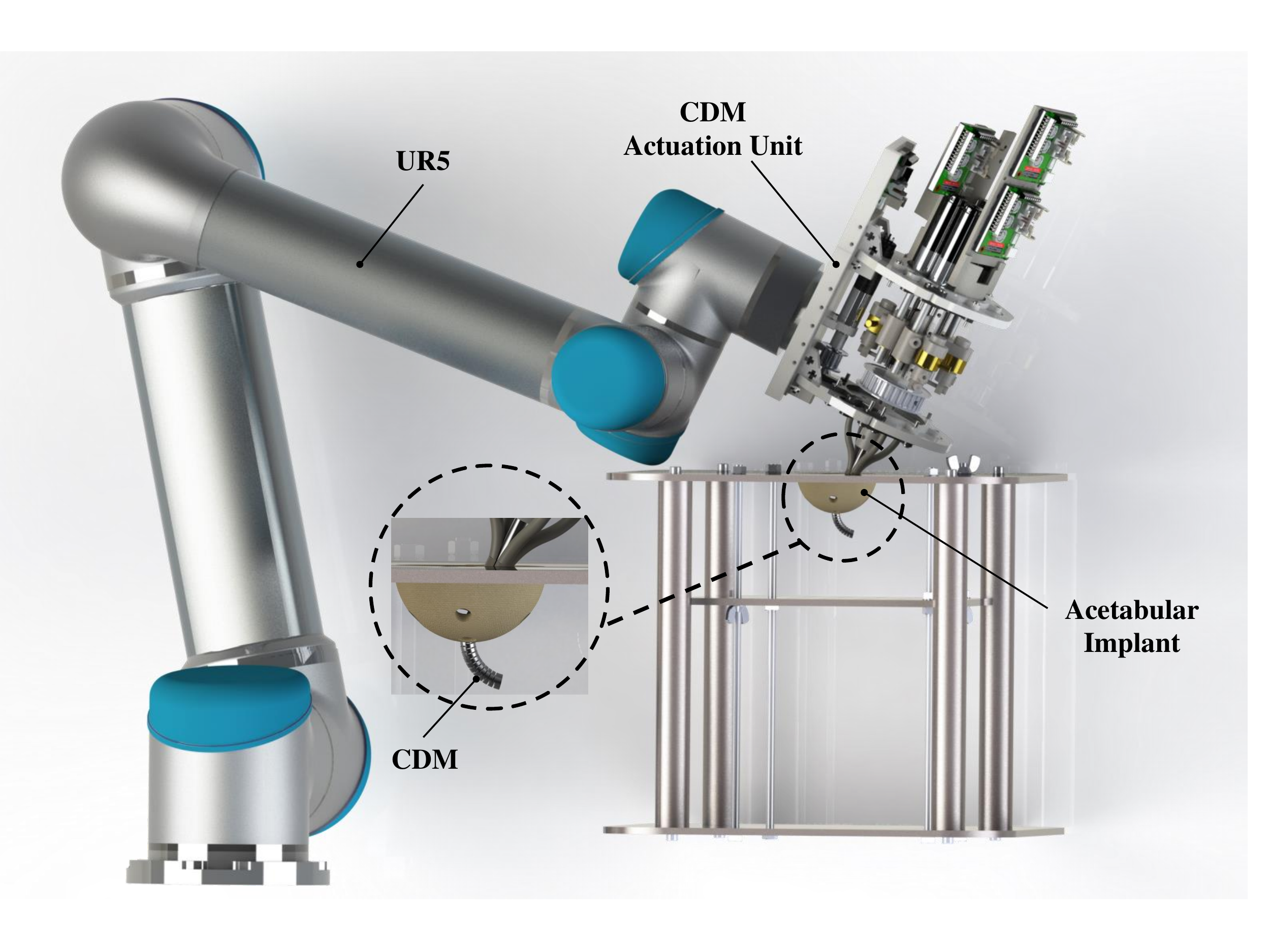}
	\caption{Hybrid redundant surgical system for treatment of osteolysis.}
	\label{fig:background}
\end{figure}

Redundancy resolution of manipulators has been widely studied in the literature. For instance, Chirikjain and Burdick \cite{chirikjian1995kinematics} and Sen et al. \cite{sen2003variational} used variational calculus for motion control of hyper-redundant manipulators and parallel robots, respectively. Further, other researchers implemented pseudoinverse control approach for dexterity optimization \cite{nakamura1987optimal} and torque minimization \cite{hollerbach1987redundancy}. Kapoor et al. \cite{kapoor2005suturing} has investigated redundancy resolution of a hybrid surgical system with a CDM using an optimization framework proposed by Funda et al. \cite{funda1993optimal}. In \cite{alambeigi2014control} and \cite{wilkening2017development}, we also incorporated a similar approach to control our proposed surgical system. In another study, Bajo et al. \cite{bajo2012integration} used the pseudoinverse-based method to control their surgical subsystem consisting of planar parallel mechanisms and continuum snake-like arms. These methods mostly use optimization-based approaches to solve a constraint-free or constrained redundancy resolution problem. However, to the best of our knowledge, this problem has not yet been studied from a convex optimization perspective. Therefore, in this paper, due to the convex nature of our problem formulation (i.e. constrained $\ell_2$-regularized quadratic minimization), we are particularly interested in convex optimization approaches, which are fast enough and easy to implement for online control of our hybrid redundant system. To this end, we choose an Alternating Direction Method of Multipliers (ADMM) algorithm \cite{boyd2011distributed}, which has been widely used in the literature due to its less complicated computations that can be solved in a parallel and distributed processing. 
% Some examples of the ADMM applications include real-time model predictive control (MPC) \cite{khadem2016ultrasound} and image processing \cite{afonso2011augmented}.

The ADMM algorithm is related to the augmented Lagrangian method (ALM) and was developed by Glowinski and Marrocco and Gabay and Mercier in mid-1970s \cite{boyd2011distributed}. It combines the superior convergence properties of the method of multipliers and the decomposability of dual ascent \cite{bertsekas2015convex}. Motivated by the various applications of the ADMM algorithm particularly in solving real-time MPC control problems \cite{khadem2016ultrasound}, in this paper, we investigate the potential implementation of this algorithm for redundancy resolution of robotic manipulators. To this end, we study the convergence conditions of this algorithm in the case of motion control of redundant hybrid robots subject to linear equality and inequality constraints. Further, as a case study, we evaluate performance of this algorithm in motion control of our surgical system via simulation.

\section{constrained motion control of redundant manipulators using the DLM formulation}
The basic equation defining forward kinematics of robotic manipulators, which relates the joint space velocities ${\dot{\theta}} \in \mathbb{R}^{n}$   to the end-effector velocity $\dot{x} \in \mathbb{R}^{m}$,  is written as follows:
%<*eq01>
\begin{equation}
\label{eq:equation01}
\dot{x} ={ J(\theta)}\dot{\theta}
\end{equation}
%</eq01>
where ${J(\theta)}\in \mathbb{R}^{m \times n}$ is the Jacobian matrix and is a function of manipulator joint angles. If we can calculate the Jacobian matrix in each time instance, then using (\ref{eq:equation01}) the changes in the end-effector position $\Delta\mathbf{x}$ for an infinitesimal time period can be estimated based on joint angles changes  $\Delta\mathbf{\theta}$:
%<*eq02>
\begin{equation}
\label{eq:equation02}
\Delta x = J(\theta)\Delta{\theta}
\end{equation}
%</eq02>
Solving this equation is dependent on the matrix $J$ and generally does not provide unique solutions. 
%For a full rank (non-singular configurations) square (non-redundant manipulators) Jacobian matrix, equation (\ref{eq:equation02}) returns a unique solution.
 For redundant $(m<n)$ manipulators, infinite solutions $\Delta\theta{}$ exist. Several methods, such as the Jacobian transpose method, pseudo inverse method, and damped least square method (DLM)  have been proposed in literature for solving (\ref{eq:equation02}). The Jacobian transpose and pseudo inverse methods encounter problems in the configurations where  ${J(\theta)}$  becomes either rank-deficient or ill-conditioned. However, DLM does not encounter the singularity problems and can provide a numerically stable procedure for calculating $\Delta\mathbf{\theta}$. In this approach, instead of solving for $\Delta\mathbf{\theta}$ satisfying (\ref{eq:equation02}), the following convex least-square optimization problem is solved:
%<*eq03>
\begin{equation}
\label{eq:equation03}
\operatornamewithlimits{argmin}_{\Delta \theta} \| J(\theta) \Delta \theta - \Delta x\|_2^2 + \lambda \| \Delta \theta \|_2^2
\end{equation}
%</eq03>
where  ${J(\theta)}\in \mathbb{R}^{m \times n}$  $(m\le n)$ and ${\lambda\in \mathbb{R}>0}$ is a positive non-zero damping constant. This optimization problem tries to find a set of $\Delta\mathbf{\theta}$ that minimizes the error between the desired task-space displacement and the generated task-space movement by the robot (first term), considering the feasible minimum joint-space displacement (second term).

 In the case of an unconstrained environment, (\ref{eq:equation03}) has closed form solution for  redundancy resolution problems. However, most of the real-world problems impose constraints on the movement of the robot. 
Therefore, we modify (\ref{eq:equation03}) to formulate the constrained redundancy resolution problem under linear inequality constraints as the following:
%<*eq04>
\begin{align}
\label{eq:equation04}
\operatornamewithlimits{argmin}_{\Delta \theta} \| J(\theta) \Delta \theta - \Delta x\|_2^2 + &\lambda \| \Delta \theta \|_2^2 \\
\text{subject to} \quad A \Delta \theta \leq b \nonumber
\end{align}
%</eq04>
where ${A(J(\theta))}\in \mathbb{R}^{r \times n}$  together with vector ${b(\theta)}\in \mathbb{R}^{r}$ define $r$ linear inequality constraints. Of note, here matrix $A$ is dependent on the Jacobian of the robotic manipulator in each time instant.

The defined constrained DLM problem in (\ref{eq:equation04}) is indeed a constrained $\ell_2$-regularized quadratic minimization problem. While this problem can be solved using various approaches of convex and non-convex optimization techniques, Ghadimi et al. \cite{ghadimi2015optimal} have shown optimally tuned ADMM method can significantly outperform existing alternatives in the literature.

\section{the ADMM algorithm \cite{boyd2011distributed}} 
The ADMM algorithm solves problems in the following form:
\begin{align}
\label{eq:equation05}
\text{minimize} \quad f(x) + g(z) \\
\text{subject to} \quad Ax+Bz = c \nonumber
\end{align}

where $f$ and $g$ are convex functions,  ${x}\in \mathbb{R}^{n}$, 〖${z}\in \mathbb{R}^{m}$, ${A}\in \mathbb{R}^{p \times n}$  , ${B}\in \mathbb{R}^{p \times m}$, and ${c}\in \mathbb{R}^{p}$. To solve this problem, we need to first form the Augmented Lagrangian (AL):
%<*eq06>
\begin{multline}
\label{eq:equation06}
L_\rho(x,z,y) = f(x) + g(z) + \\ y^{T} (Ax + Bz -c ) + (\frac{\rho}{2}) \| Ax + Bz - c \|_2^2
\end{multline}
%</eq06>
where ${y}\in \mathbb{R}^{p}$ is a vector of Lagrange multipliers and $\rho\ge 0$ is a constant called the AL penalty parameter. In the ADMM algorithm, a new iteration $(x_{k+1},z_{k+1},y_{k+1})$,  is generated given the current iteration $(x_{k},z_{k},y_{k})$ by first minimizing the AL with respect to $x$, then with respect to $z$, and finally updating the multiplier $y$:
%<*eq07>
\begin{align}
\label{eq:equation07}
%\begin{split}
x_{k+1} &= \operatornamewithlimits{argmin}_x {} f(x) +  y_k^T (Ax + B (\frac{\rho}{2}) \| Ax + Bz_k - c \|_2^2  \\ \nonumber
%\end{split} \\
%\begin{split}
z_{k+1} &= \operatornamewithlimits{argmin}_z {} g(z) + y_k^T (Ax_{k+1} -(\frac{\rho}{2}) \| Ax_{k+1} + Bz - c \|_2^2 \\ \nonumber
%\end{split} \nonumber \\
y_{k+1} &= y_k + \rho (A x_{k+1} + B z_{k+1} - c) 
\end{align}
%</eq07>

In this paper, we will use the scaled form of the ADMM algorithm by combining the linear and quadratic terms in the AL, and scaling the dual variable $ u=y \slash {\rho}$ (Algorithm \ref{Algorithm ADMM}).
\begin{algorithm} \label{Algorithm ADMM}
	%     \SetKwInOut{Input}{Input}
	%     \SetKwInOut{Output}{Output}    
	%     \Input{Jacobian at time step k, $J_k$ \\ 
	%     Target tip position, $x_{target}$ \\
	%     Threshold, $\epsilon$} 
	%     \Output{Actuation input $\Delta l_k$ \\ 
	%     Jacobian at the next step}
	Choose $(z_0, u_0)$, $\rho > 0$ and set $k \leftarrow 0$ \\
	\textbf{start loop} \\
	Compute (in order) the following updates: \\
	$x_{k+1} = \operatornamewithlimits{argmin}_x {} f(x) + (\frac{\rho}{2}) \| Ax + Bz_k - c + u_k\|_2^2$  \\
	$z_{k+1} = \operatornamewithlimits{argmin}_z {} g(z) + (\frac{\rho}{2}) \| Ax_{k+1} + Bz - c + u_k\|_2^2$ \\
	$u_{k+1} = u_k + \rho (A x_{k+1} + B z_{k+1} - c)$ \\
	set $k \leftarrow k+1 $ \\
	\textbf{end loop} some stopping criterion will be satisfied.
	\caption{ADMM}
\end{algorithm}

Convergence of the ADMM algorithm is usually defined by the primal $r_{k+1}$ and dual $s_{k+1}$  residuals:
\begin{align}
\label{eq:equation08}
r_{k+1} &= (A x_{k+1} + B z_{k+1} - c) \\
s_{k+1} &= \rho A^T B(z_{k+1} -z_{k}) \nonumber
\end{align}
Using these residuals, a termination criterion is defined as: 
\begin{align}
\label{eq:equation09}
\left\| r_{k}\right\|_{2} \le \epsilon_{pri} \quad \text{and} \quad  \left\| s_{k}\right\|_{2} \le \epsilon_{dual}
\end{align}
where $\epsilon_{pri} > 0$ and $\epsilon_{dual} > 0$ are positive feasibility tolerances for the primal and dual feasibility conditions. These tolerances can be defined as follows:
\begin{align}
\label{eq:equation10}
\epsilon_{pri} &=\sqrt{p}\; \epsilon_{abs}+\epsilon_{rel} \; max\{\left\|Ax_{k}\right\|_{2},\left\|Bz_{k}\right\|_{2},\left\|c\right\|_{2}\} \\
\epsilon_{dual} &=\sqrt{n}\; \epsilon_{abs}+\epsilon_{rel} \;\left\|\rho A^T u\right\|_{2} \nonumber
\end{align}
where $\epsilon_{abs}> 0$ and $\epsilon_{rel}> 0$ are absolute and relative tolerances, and $\sqrt{p}$ and $\sqrt{n}$ are coefficients demonstrating $\ell_2$ norms in $ \mathbb{R}^{p}$ and $\mathbb{R}^{n}$, respectively.
%Investigation of the scaled ADMM algorithm demonstrates that it is only dependent on the AL penalty parameter $\rho$. It is proved that convergence of this algorithm under some assumptions is independent of this parameter. However, this parameter has a direct effect on the convergence rate of the algorithm \cite{ghadimi2015optimal}. 

\section{ADMM algorithm for the constrained DLM problem of redundant robots}

To apply the ADMM algorithm to the defined constrained DLM problem  (\ref{eq:equation04}), we first convert it to the standard ADMM form (\ref{eq:equation05}). To this end, we use a slack vector $z \ge0$ and put an infinite penalty on its negative components. Considering this, we can rewrite (\ref{eq:equation04}) as follows:
%<*eq04>
\begin{align}
\label{eq:equation11}
&\operatornamewithlimits{argmin}_{\Delta \theta} \| J(\theta) \Delta \theta - \Delta x\|_2^2 + \lambda \| \Delta \theta \|_2^2 + I_+(z)\\
&\text{subject to} \quad A \Delta \theta-b+z=0 \nonumber
\end{align}
%</eq04>
where  ${J(\theta)}\in \mathbb{R}^{m \times n}$  $(m\le n)$, $ {\Delta{\theta}} \in \mathbb{R}^{n}$,  $\Delta{x} \in \mathbb{R}^{m}$,  and ${\lambda\in \mathbb{R}>0}$.  ${A(J(\theta))}\in \mathbb{R}^{r \times n}$  together with vector ${b(\theta)}\in \mathbb{R}^{r}$ define $r$ linear inequality constraints.  $I_+ (z)$ is the indicator function of the positive orthant defined as $I_+ (z)=0$ for $z \ge 0$ and $I_+ (z)=\infty$ otherwise. The associated augmented Lagrangian of (\ref{eq:equation11}) in the scaled form would be:
\begin{multline}
\nonumber
L_\rho(\Delta{\theta},z,y) = \| J(\theta) \Delta \theta - \Delta x\|_2^2 +\lambda \| \Delta \theta \|_2^2 + I_+(z)+\\ (\frac{\rho}{2}) \| A\Delta{\theta}
+ z - b+u \|_2^2
\end{multline}
which leads to the following scaled ADMM iterations:
\begin{align}
\label{eq:equation12}
\Delta{\theta}_{k+1} &=(J^TJ+\lambda I_n+ \rho A^TA)^{-1} (J^T\Delta{x}-\rho A^T(z_k-b+u_k)) \nonumber \\
z_{k+1} &=max\{0,(-A\Delta{\theta_{k+1}+b-u_k})\} \\ 
u_{k+1} &=u_k+A\Delta{\theta}_{k+1}-b+z_{k+1} \nonumber
\end{align}
where $I_n$ is the $n \times n$ identity matrix. 

\subsection{Proof of Convergence}
Inspired by the approach proposed in \cite{ghadimi2015optimal}, in this section, we show that the convergence of  the constrained DLM problem (\ref{eq:equation11}) under some conditions is independent of the choice of $\rho>0$, which makes this algorithm a powerful method in solving constraint redundancy resolution problems of robots. % Further, we introduce an optimal AL penalty parameter, which results in the fastest convergence under some conditions.
 To prove the convergence of (\ref{eq:equation11}), a vector of indicator variables $d_k^n\{0,1\}$ is defined to check whether the $i^{th}$ inequality constraint in (\ref{eq:equation11}) is active or not. In this definition, $d_k^i=1$ means the  $i^{th}$ inequality constraint is active (i.e. $u_k^i\ne 0$ in \ref{eq:equation12}) or the slack variable $z_i $ equals zero at the current iteration. Considering this, we introduce the following auxiliary variables:
\begin{align}
\label{eq:equation13}
\upsilon_k &= z_k+u_k, \quad \upsilon_k {\in \mathbb{R}>0}\\
D_k &= diag (d_k) \nonumber
\end{align}
From the definition of $D_k$ and $\upsilon_k$, we have $D_k \upsilon_k=u_k$ and $(I-D_k)\upsilon_k=z_k.$ Now, we can rewrite $z$- and $u$-updates of (\ref{eq:equation12}) as:
\begin{align}
\label{eq:equation14}
\upsilon_{k+1} &= H_{k+1}(D_k\upsilon_k+A\Delta{\theta_{k+1}}-b)\\
H_{k+1} &= diag(sign(D_k\upsilon_k+A\Delta{\theta_{k+1}}-b))  \nonumber
\end{align}
where $sign()$ function returns the sign of the elements of its vector argument. Therefore, using (\ref{eq:equation14}) we can rewrite (\ref{eq:equation12}) as follows:
\begin{align}
\label{eq:equation15}
\Delta{\theta}_{k+1} &= (J^TJ+\lambda I_n+ \rho A^TA)^{-1} (J^T\Delta{x}-\rho A^T(z_k-b+u_k)) \nonumber \\
\upsilon_{k+1} &= H_{k+1}(D_k\upsilon_k+A\Delta{\theta_{k+1}}-b) \\
D_{k+1} &= 1/2 (I+H_{k+1}) \nonumber
\end{align}
where $D_{k+1}$ updates are derived from the definition of $D_{k+1}$ and $H_{k+1}$. In (\ref{eq:equation15}), we use the definition of $\upsilon_k$ and substitute $\Delta{\theta}_{k+1}$ in $\upsilon_{k+1}$ to obtain:
\begin{align}
\label{eq:equation16}
\upsilon_{k+1} = H_{k+1}(D_k-A(J^TJ+\lambda I_n+ \rho A^TA)^{-1}\rho A^T)\upsilon_{k}\\ \nonumber
-H_{k+1}(A(J^TJ+\lambda I_n+ \rho A^TA)^{-1}(-J^T \Delta{x} \rho A^Tb)+b)
\end{align}
Let’s define matrix $P$ as
\begin{align}
\label{eq:equation17}
P=A(J^TJ+\lambda I_n+ \rho A^TA)^{-1}\rho A^T
\end{align}
So, using (\ref{eq:equation16}) we can obtain
\begin{align}
\label{eq:equation18}
H_{k+1}\upsilon_{k+1}-H_{k}\upsilon_{k}=(I/2-P)(\upsilon_{k}-\upsilon_{k-1})\\
+1/2(H_k\upsilon_{k}-H_{k-1}\upsilon_{k-1}) \nonumber
\end{align}
We use (\ref{eq:equation18}) to prove the convergence of the ADMM algorithm and find the optimal penalty parameter $\rho$ that results in the fastest convergence rate. 

\begin{theorem}
	The constraint redundancy resolution problem (\ref{eq:equation11}) with the ADMM iterations (\ref{eq:equation12}) converges to zero at linear rate for all the AL penalty parameter ρ if:
	\begin{enumerate}
		\item $\rho$ is a real positive value $({\rho\in \mathbb{R}>0})$. 
		\item damping constant $\lambda$  is a real positive value $({\lambda\in \mathbb{R}>0})$. 	
	\end{enumerate}
\end{theorem}
\textit{proof:} We start from (\ref{eq:equation18}) and take the norm of both sides
\begin{align}
\label{eq:equation19}
\|H_{k+1}\upsilon_{k+1}-H_{k}\upsilon_{k}\|\le 1/2\|(I-2P)\|\|(\upsilon_{k}-\upsilon_{k-1})\|\\
+1/2\|(H_k\upsilon_{k}-H_{k-1}\upsilon_{k-1})\| \nonumber
\end{align}
Since $H_k$ is a diagonal matrix with $\pm {1}$ elements and $\upsilon_{k}$ is a positive vector, we can write:
\begin{align}
\|(\upsilon_{k}-\upsilon_{k-1})\|\le\|(H_k\upsilon_{k}-H_{k-1}\upsilon_{k-1})\|
\end{align}
Therefore, we can rewrite (\ref{eq:equation19}) as
\begin{align}
\label{eq:equation20}
\|H_{k+1}\upsilon_{k+1}-H_{k}\upsilon_{k}\|\le \underbrace{{1/2(\|(I-2P)\|+1)\|}}_{\gamma}.\\
\|(H_k\upsilon_{k}-H_{k-1}\upsilon_{k-1})\| \nonumber
\end{align}
In (\ref{eq:equation20}), linear convergence to zero is guaranteed for $\gamma<1$, which implies $\|(I-2P)\|<1$. Further, considering the definition of $P$  in (\ref{eq:equation17}), the convergence rate depends on the parameters $\lambda$ and $\rho$.

To prove linear convergence of (\ref{eq:equation20}), we first need to calculate$\|(I-2P)\|$. To this end, we apply the matrix inversion lemma on the modified version of (\ref{eq:equation17}):
\begin{align}
P=A(\underbrace{(J^TJ+\lambda I_n)}_{Q}/\rho+ A^TA)^{-1} A^T\nonumber
\end{align}
By the matrix inversion lemma, we have
\begin{align}
\label{eq:equation21}
P=\rho A Q^{-1}A^T-\rho A Q^{-1}A^T(I+\rho A Q^{-1}A^T)\rho A Q^{-1}A^T
\end{align}
In (\ref{eq:equation21}), calculation of $P$ depends on the invertibility of matrix $Q$ and $(I+\rho A Q^{-1}A^T)$. For all $\lambda>0$, adding $\lambda I_n$  to the positive semidefinite matrix $J^T J$ makes $Q$ a positive definite matrix. Therefore, it is invertible and all its eigenvalues are real and positive. Since  $Q$ is a positive definite matrix, $A Q^{-1}A^T$ is a positive semi definite matrix for any choice of matrix $A$. Further, with similar analogy, $(I+\rho A Q^{-1}A^T)$ is a positive definite matrix and invertible for all $\rho>0$.

To calculate $\gamma$  in (\ref{eq:equation20}), we now use the following relation and its dependency to eigenvalues of $P$ \cite{horn2012matrix}:
\begin{align}
\label{eq:equation22}
\gamma=1/2\|(I-2P)\|+1/2=\underset{i}{\text{max}} (1/2|(1-2\sigma_i(P)|+1/2)
\end{align}
where $\sigma_i(P)$ are the eigenvalues of matrix $P$. Considering the fact that $(I+\rho A Q^{-1}A^T)$ is a polynomial function of $t=\rho A Q^{-1}A^T$, eigenvalues of $P$ in (\ref{eq:equation21}) can be calculated as the following \cite{horn2012matrix}:
\begin{align}
P=\Phi(t)\Rightarrow P=t-t(1+t)^{-1}t \nonumber
\end{align}
We calculate eigenvalues of $P$ (i.e. $\sigma_i(P)$) as a function of eigenvalues of $\rho A Q^{-1}A^T$  (i.e. $\sigma_i(\rho A Q^{-1}A^T)$) \cite{horn2012matrix}:
\begin{align}
\label{eq:equation23}
\sigma_i(P)=\frac{\sigma_i(\rho A Q^{-1}A^T)}{1+\sigma_i(\rho A Q^{-1}A^T)}
\end{align}
In this equation $\rho A Q^{-1}A^T)$ is a positive semi-definite matrix which, based on (\ref{eq:equation22}), implies $\sigma_i(P) \in [0,1) \Rightarrow \|(I-2P)\|\le1 $. Therefore, if $\rho>0$ and 

\textit{Case 1:} $A$ is invertible or is a full row-rank matrix then $\sigma_i(P) \in (0,1) \Rightarrow \|(I-2P)\|<1 $, which proves the linear convergence of (\ref{eq:equation20}) to zero.

\textit{Case 2:} $A$ is a tall matrix with full column rank (i.e. $A$ is not a full row-rank matrix) then the case of  $\sigma_i(P)=0 \Rightarrow \|(I-2P)\|=1 $ might arise. To check the linear convergence of (\ref{eq:equation21}) in this case, from (\ref{eq:equation15}) we can write:
\begin{align}
\Delta{\theta}_{k+1}-\Delta{\theta}_{k}=-(J^TJ+\lambda I_n+ \rho A^TA)^{-1}\rho A^T(\upsilon_{k}-\upsilon_{k-1}) \nonumber
\end{align}
Considering the definition of $P$ in (\ref{eq:equation17}), we multiply both sides by $A$ to obtain:
\begin{align}
A\Delta{\theta}_{k+1}-\Delta{\theta}_{k}=-P(\upsilon_{k}-\upsilon_{k-1}) \nonumber
\end{align}
Given $\sigma_i(P)=0$ in \textit{Case 2}, a vector 〖$(\upsilon_{k}-\upsilon_{k-1})\neq 0$ exists in the null space of $P$, which implies  $A\Delta{\theta}_{k+1}-\Delta{\theta}_{k}=0$. This means either vector $\Delta{\theta}_{k+1}-\Delta{\theta}_{k}$ is in the null space of $A$ or 〖$\Delta{\theta}_{k+1}=\Delta{\theta}_{k}$. Considering assumptions in \textit{Case 2}, $A$ is a full column rank matrix implying vector $\Delta{\theta}_{k+1}-\Delta{\theta}_{k}$ is not in the null space of matrix $A$. In other words, 〖$\Delta{\theta}_{k+1}=\Delta{\theta}_{k}$  are the stationary points of the algorithm (\ref{eq:equation12}). Therefore, algorithm (\ref{eq:equation12}) converges linearly to zero and the case of zero eigenvalues for matrix  $P$ can be neglected. $\square$
\subsection{Optimal Convergence Parameters } \label{convergence parameter}
In the previous section, we proved that for all $\rho>0$, the ADMM iterations in (\ref{eq:equation12}) linearly converge to zero. To obtain the fastest convergence rate, we use the following optimal AL penalty parameter $\rho^*$  introduced by Ghadimi et al. \cite{ghadimi2015optimal} that  results in the fastest convergence rate assuming all the conditions of \textit{Theorem 1} hold and the constraint matrix $A$, in the redundancy resolution problem (\ref{eq:equation11}) and the corresponding ADMM iteration (\ref{eq:equation12}), is either full row-rank or invertible:
\begin{align}
\label{eq:equation25}
\rho^*=\frac{1}{\sqrt{ \sigma_{min}(A Q^{-1}A^T)\sigma_{max}(A Q^{-1}A^T)}}
\end{align}
where $\sigma_{min}$  and  $\sigma_{max}$ are the minimum and maximum eigenvalues of $A Q^{-1}A^T$, respectively. Furthermore, when rows of $A$ are linearly dependent, $\rho^*$ can still reduce the convergence time if $\sigma_{min}$ is replaced by the smallest nonzero eigenvalue of $A Q^{-1}A^T$.
\section{case study: a redundant surgical system for treatment of osteoyisis}
We are developing an MIS  robotic system to remove and treat osteolytic lesions behind a well-fixed implant (Fig. \ref{fig:background}). This surgical workstation consists of a six DoF robotic arm (UR5, Universal Robotics- Denmark) integrated with a one DoF cable-driven continuum dexterous manipulator (CDM), which are concurrently controlled to position the CDM tip behind the acetabular implant \cite{alambeigi2014control},\cite{alambeigi2016design}. Position of the CDM tip behind the acetabular implant is controlled using concurrent control of the coupled CDM-robotic manipulator. 
In this paper, we use this redundant system to evaluate performance of the proposed ADMM algorithm. Furthermore, we define appropriate linear constraints to satisfy both operational and safety objectives, which are necessary during robot-assisted treatment of osteolysis. To this end, considering the DLM formulation introduced in (\ref{eq:equation11}), kinematics of the redundant robot as well as the appropriate linear constraints will completely be defined in the following sections. 
\subsection{Kinematics of the integrated system}
\subsubsection{Kinematics of the UR5} 
\begin{figure}[!t]
	\centering
	\includegraphics[width=\linewidth]{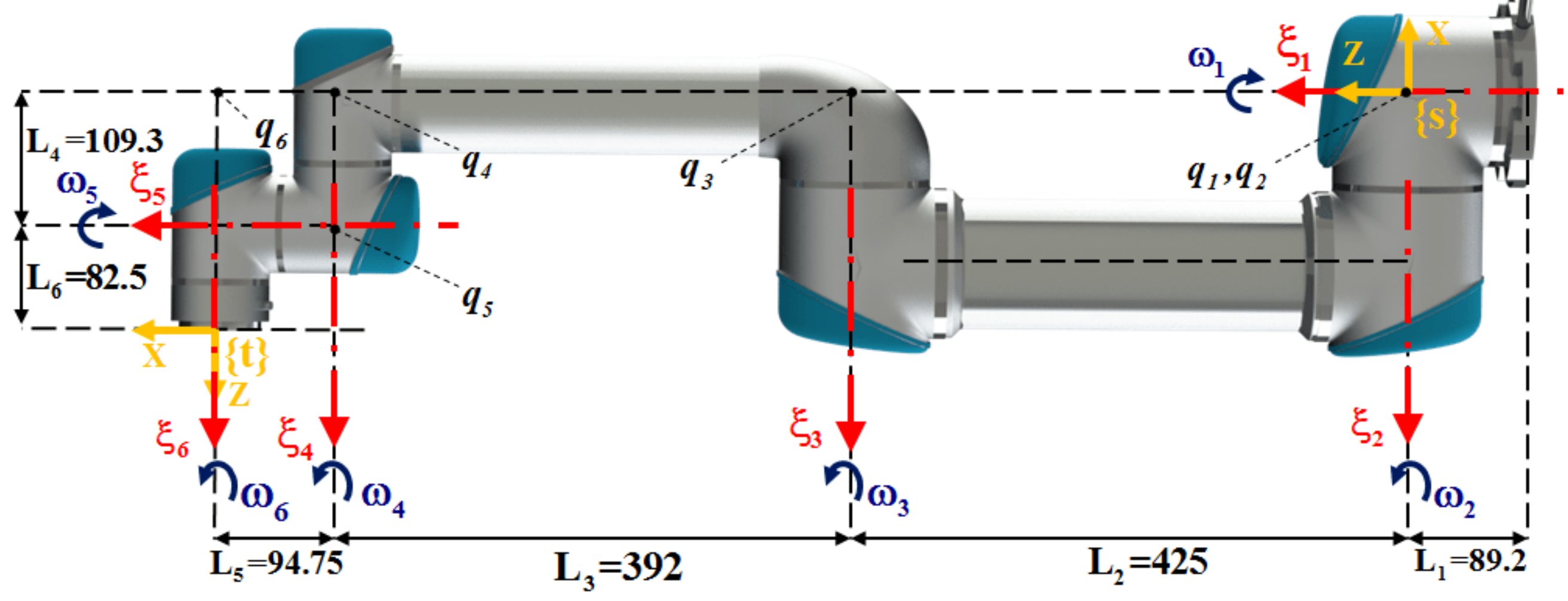}
	\caption{Considered twist axes $(\zeta_i)$, angular velocities $(\omega_i)$ and points on axes $(q_i)$ in finding forward kinematics of the UR5.}
	\label{fig:UR5}
\end{figure}

The forward kinematics of a manipulator is obtained by a mapping $g_{st}: \Theta \rightarrow SE(3)$, which describes the end-effector configuration as a function of the robot joint variables $\Theta $ and can be written using the product of exponentials formula:
\begin{align}
\label{eq:equation26}
g_{st} &= e^{\hat{\zeta_1}\Theta_1}e^{\hat{\zeta_2}\Theta_2}...e^{\hat{\zeta_n}\Theta_n}g_{st}(0) \\
\hat{\zeta} &=
\begin{bmatrix}
\hat{\omega}       & -\omega \times q \\
0       & 0  \\
\end{bmatrix},
\hat{\omega} \in so(3), q \in \mathbb{R}^{3},\Theta \in \mathbb{R}^{n} \nonumber
\end{align}
where $\omega$ is the axis of rotation, $q$ is a point on the axis and $\zeta=[-\omega \times q \quad \omega ]^T$ is the twist corresponding to the $i^{th}$ joint axis in the reference configuration \cite{murray1994mathematical}. Fig. \ref{fig:UR5} demonstrates the twist axes $\zeta$ for all six joints of UR5. 
%Considering this figure, we can compute the twists of all the joints and $g_{st} (0)$- as the base to tip transformation in the reference configuration- as follows:
%\begin{align}
%\nonumber
%g_{st}(0)=
%\begin{bmatrix}
%0       &0  &-1  &-L_4-L_6\\
%0       &1 &0 &0  \\
%1       &0  &0 &L_2+L_3+L_5\\
%0  &0 &0 &1 
%\end{bmatrix}
%\end{align}
The velocity of the base of the CDM relative to the base of UR5 $(\dot{x}_{CDM \  base}^{UR5 \ base})$ in the Cartesian space is related to the UR5 joint velocities  $(\dot{\theta}_{UR5} \in\mathbb{R}^{6})$ via its Jacobian matrix $J_{UR5}\in\mathbb{R}^{6\times 6}$ as follows:
%<*eq28>
\begin{equation}
\label{eq:equation27}
\dot{x}_{CDM \ base}^{UR5 \ base} = J_{UR5} \dot{\theta}_{UR5}
\end{equation}
%</eq28>
where the manipulator Jacobian $J_{UR5}$   has the following form:
%<*eq29>
\begin{equation}
\label{eq:equation28}
\nonumber
J_{UR5}(\theta) = [\zeta_1 \ \zeta_2^{\prime} \ ... \ \zeta_6^{\prime}];   \\
\quad \zeta_i^{\prime} = Ad_{\exp^{\hat{\zeta_1} \theta_1} \ ... \ \exp^{\hat{\zeta_{i-1}} \theta_{i-1}}}\zeta_i
\end{equation}
%</eq29>
where $Ad_ {(\ ... )}\in \mathbb{R}^{6\times 6}$ is the adjoint matrix, which depends on the UR5 configuration \cite{murray1994mathematical}.
\subsubsection{Kinematics of the CDM \cite{alambeigi2014control}}
The 35 mm CDM is made of two nested nitinol tubes with notches on its surface and has outer diameter of 6 mm and inner diameter of 4 mm. Considering these notches, two actuating cables on the sides of the CDM provide a planar bend for the continuum manipulator. The inner space of the CDM is used as the tool channel for passing different types of instruments used for the debriding and treatment process. Two DC motors (RE16, Maxon Motor Inc.) with spindle drives (GP16, Maxon Motor, Inc.) actuate the CDM via the actuation cables (Fig. 1).
In \cite{alambeigi2014control}, we have derived experimental kinematics of the CDM in the free bending motion. A series of experimental tests have been performed to investigate the relation between the actuating cable length $(l)$ and tip position of the CDM $(p)$. Using this experimental function, the velocity of the CDM tip relative to its base $\dot{x}_{CDM \ base}^{UR5 \ base}$   can be determined by the following partial differentiation: 
%<*eq31>
\begin{equation}
\label{eq:equation29}
\dot{x}_{CDM \ tip}^{UR5 \ base} = \frac{\partial p}{\partial l} \cdot \frac{\partial l}{\partial t}
\end{equation}
%</eq31>
\subsubsection{	Combined Kinematics}
The integrated system has 7 DoF $(\dot{\theta}_{combined} \in \mathbb{R}^{7\times 1})$, six for the UR5 and one for the CDM. Therefore, the combined Jacobian of the CDM and UR5 $(J_{combined} \in \mathbb{R}^{6\times 7})$ can be calculated using (\ref{eq:equation27}) and (\ref{eq:equation29}) as follows:
%<*eq32>
\begin{align}
\label{eq:equation32}
J_{combined} &= [J_{UR5} \quad J_{CDM}] \nonumber \\
J_{UR5} &\in \mathbb{R}^{6 \times 6} \nonumber \\ 
J_{CDM} &\in \mathbb{R}^{6 \times 1} \nonumber \\
\dot{x}_{CDM \ tip}^{UR5 \ base} &= J_{combined} \dot{\theta}_{combined}
\end{align}
%</eq32>
\subsection{Constraints} \label{constraints}
\begin{figure}[!t]
	\centering
	\includegraphics[width=\linewidth]{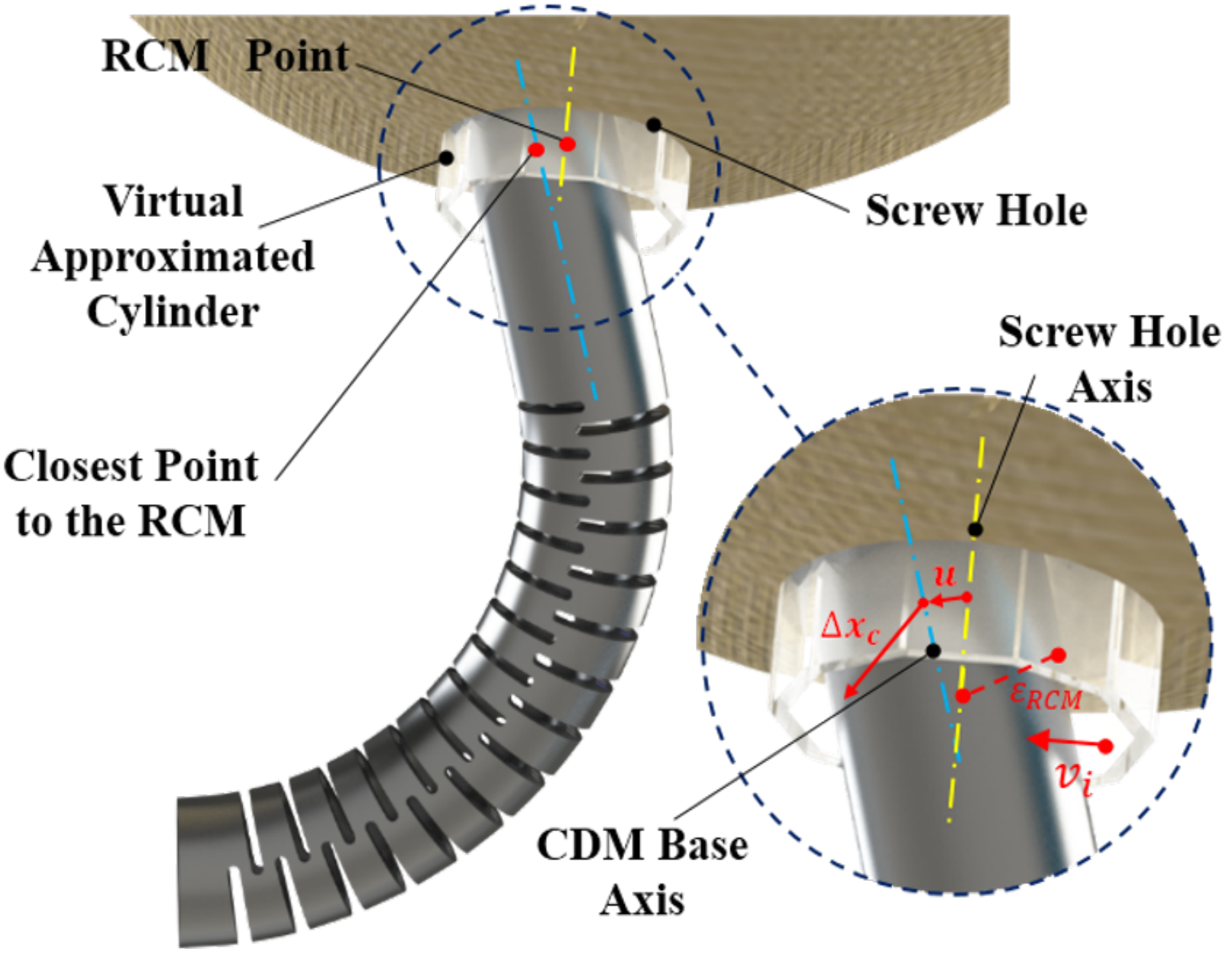}
	\caption{Defining the RCM constraint based on the approximated virtual cylinder with a polygon.}
	\label{fig:RCM}
\end{figure}
\subsubsection{The Remote Center of Motion (RCM) constraint}
For robotic manipulators without a mechanical RCM (e.g., the UR5), a virtual RCM can be applied \cite{alambeigi2014control}. This constraint ensures movements of the CDM base are confined in a virtual cylinder around the screw hole axis with a radius of  $\epsilon_{RCM}$ (Fig.\ref{fig:RCM}). Further, it prevents any collision between the CDM base and the screw hole edges while the integrated system pivots around the center of the hole with the radius $r$. We can define this virtual constraint as:
%<*eq33>
\begin{align}
\label{eq:equation33}
\underbrace{
	\begin{bmatrix}
	v_1 \\
	\vdots \\
	v_m
	\end{bmatrix}
	\cdot J_{closest \ points} }_{A_1} \cdot \Delta \theta_{UR5} &\leq 
\underbrace{
	\begin{bmatrix}
	\epsilon_{RCM} + v_1 \cdot u \\
	\vdots \\
	\epsilon_{RCM} + v_m \cdot u
	\end{bmatrix} }_{b_1} \\
\Delta x_c = J_{closest \ point} &\cdot \Delta \theta_{UR5}; \nonumber \\
A_1 \in \mathbb{R}^{m \times 7}, \Delta \theta_{UR5} &\in \mathbb{R}^{6 \times 1}
\end{align}
%</eq33>
where vectors $v_i$ are vectors normal to each side of the polygon which approximates the cylinder and $m$ defines the degree of approximation of a circle by a polygon. $\Delta{x}_c$  is the incremental Cartesian movement of the closest point on the CDM base axis to the RCM point and $J_{closest \ point}$  is the Jacobian matrix of the closest point in the UR5 base frame.
\subsubsection{Joints Limit Constraints}
To ensure a safe range for the incremental joint movements of the UR5, we defined:
%<*eq34>
\begin{align}
\label{eq:equation34}
I \cdot \Delta \theta_{UR5} &\leq \Delta \theta_{UR5 \ upper}\\ \nonumber
- I \cdot \Delta \theta_{UR5} &\leq -\Delta \theta_{UR5 \ lower}
\end{align}
%</eq34>
where $I$ is a $6 \times 6$  identity matrix and $\Delta \theta_{UR5 \ upper}$ and $\Delta \theta_{UR5 \ lower}$  refer to the lower and upper incremental joint movements of the UR5, respectively. Furthermore, considering (\ref{eq:equation27}) as the experimental model of the CDM and a single-cable bend, the following constraints are applied:
%<*eq35>
\begin{align}
\label{eq:equation35}
\Delta \theta_{CDM} &\leq 9mm - \theta_{CDM}\\ \nonumber
-\Delta \theta_{CDM} &\leq \theta_{CDM}
\end{align}
%</eq35>
This means that the total change in the cable length of the CDM ($\theta_{CDM}$), at each moment should not exceed 9 mm and clearly it should not be less than 0 mm. We can stack these joints limit constraints and write:
%<*eq36>
\begin{equation}
\label{eq:equation36}
\underbrace{
	\begin{bmatrix}
	I & 0 \\
	0 & 1 \\
	-I & 0 \\
	0 & -1
	\end{bmatrix} }_{A_2} 
\cdot \Delta \theta \leq 
\underbrace{
	\begin{bmatrix}
	\Delta \theta_{UR5 \ Upper}\\
	9mm - \theta_{CDM} \\
	-\Delta \theta_{UR5 \ lower}\\
	\theta_{CDM}
	\end{bmatrix} }_{b_2} 
\end{equation}
%</eq36>
where $\Delta{\theta} \in \mathbb{R}^{7\times 1}$, $A_2 \in \mathbb{R}^{14\times 7}$.
\subsubsection{Combining Constraints as Matrix A and Vector b}
We stack the constraints defined in (\ref{eq:equation33}) and (\ref{eq:equation36}) as matrix $A$ and vector $b$ to use them in (\ref{eq:equation11}):
%<*eq37>
\begin{equation}
\label{eq:equation37}
\underbrace{
	\begin{bmatrix}
	A_1 & 0 \\
	A_2
	\end{bmatrix} }_{A} 
\cdot \Delta \theta \leq 
\underbrace{
	\begin{bmatrix}
	b_1 \\
	b_2
	\end{bmatrix} }_{b} 
, \ \Delta \theta \in \mathbb{R}^{7 \times 1}
\end{equation}
%</eq37>
\subsection{Control Algorithm}
Considering the combined Jacobian in (\ref{eq:equation32}) and the stacked constraints in (\ref{eq:equation37}), Algorithm (\ref{control}) demonstrates the concurrent control method of the integrated UR5-CDM system. The goal of this algorithm is to use the solution of the proposed constrained DLM formulation in (\ref{eq:equation11}) and follow a pre-defined trajectory. 
% concurrent control
\begin{algorithm} \label{control}
	
	Define the desired path, the CDM tip error $\epsilon_{CDM \ tip}$ and $\epsilon_{RCM}.$\\
	\For{$k = i: size(desired path)$}{
		\While{ $\| \Delta x_{desired} \|_2 \geq \epsilon_{CDM \ tip}$}{
			Query the UR5 and the CDM current joint values and cable length ($\theta \in \mathbb{R}^{7 \times 1}$) \\
			Calculate $x^{UR5 \ base}_{CDM \ tip}$ and its distance to the desired position $x^{UR5 \ base}_{desired}$ using current joint angle: $\Delta x_{desired} = x^{UR5 \ base}_{CDM \ tip} - x^{UR5 \ base}_{desired}$ \\
			Calculate combined Jacobian, matrix $A$  and vector $b$.\\
			Compute (\ref{eq:equation11}) using (\ref{eq:equation12}): \\
			$\Delta \theta \gets \operatornamewithlimits{argmin}_{\Delta \theta} \frac{1}{2}\| J(\theta) \Delta \theta - \Delta x\|_2^2 + \frac{\lambda}{2} \| \Delta \theta \|_2^2$ \\
			Set $\theta \gets \theta_{old} + \Delta \theta$ \\
			Set $\theta_{old} \gets \theta$ \\
			Calculate $\Delta x_{desired}$ \\            
		}
		Set $k \gets k+1 $ \\
	}
	\caption{Concurrent UR5-CDM Control}
\end{algorithm}
\section{RESULTS AND DISCUSSION}
The performance of the proposed ADMM algorithm in solving the constrained concurrent control of the described surgical system is evaluated with simulation. To this end, we used a representative osteolytic lesion boundary reconstructed based on data provided by collaborating surgeons \cite{alambeigi2014control}. This complex 3D-path includes 36 waypoints and can be confined in a $7\times 7\times 7$ $cm^3$ cubic space. For this study, we assumed that the CDM already passed through one of the screw holes of the acetabular implant and  the CDM was completely inside the cavity behind the implant. In addition, considering (\ref{eq:equation28}), we assumed that no external force was acting on the CDM body during the procedure.

All simulations started with identical initial poses of the UR5 and CDM. The goal of these simulations was to track the path while satisfying all the constraints described in Section \ref{constraints} on joint limits. We approximated the RCM circle by a polygon with $m=16$ sides as described in (14). Hence, dimension of the matrix $A$ in (\ref{eq:equation33}) is  $A \in \mathbb{R}^{30\times 7}$. We used (\ref{eq:equation12}) to solve the problem with the ADMM algorithm and (\ref{eq:equation10}) as the termination criteria by setting 〖$\epsilon_{abs}=1e^{-5}$ and 〖$\epsilon_{rel}=1e^{-3}$ with $p=30$ and $n=7$. All simulations were performed in MATLAB (The MathWorks Inc, Natick, MA), on a Core i5, 2.5 GHz processor and 4 GB of RAM computer running Windows 7.
\subsection{Trajectory Tracking with Constant Parameters}
\begin{figure}[!t]
	\centering
	\includegraphics[width=\linewidth]{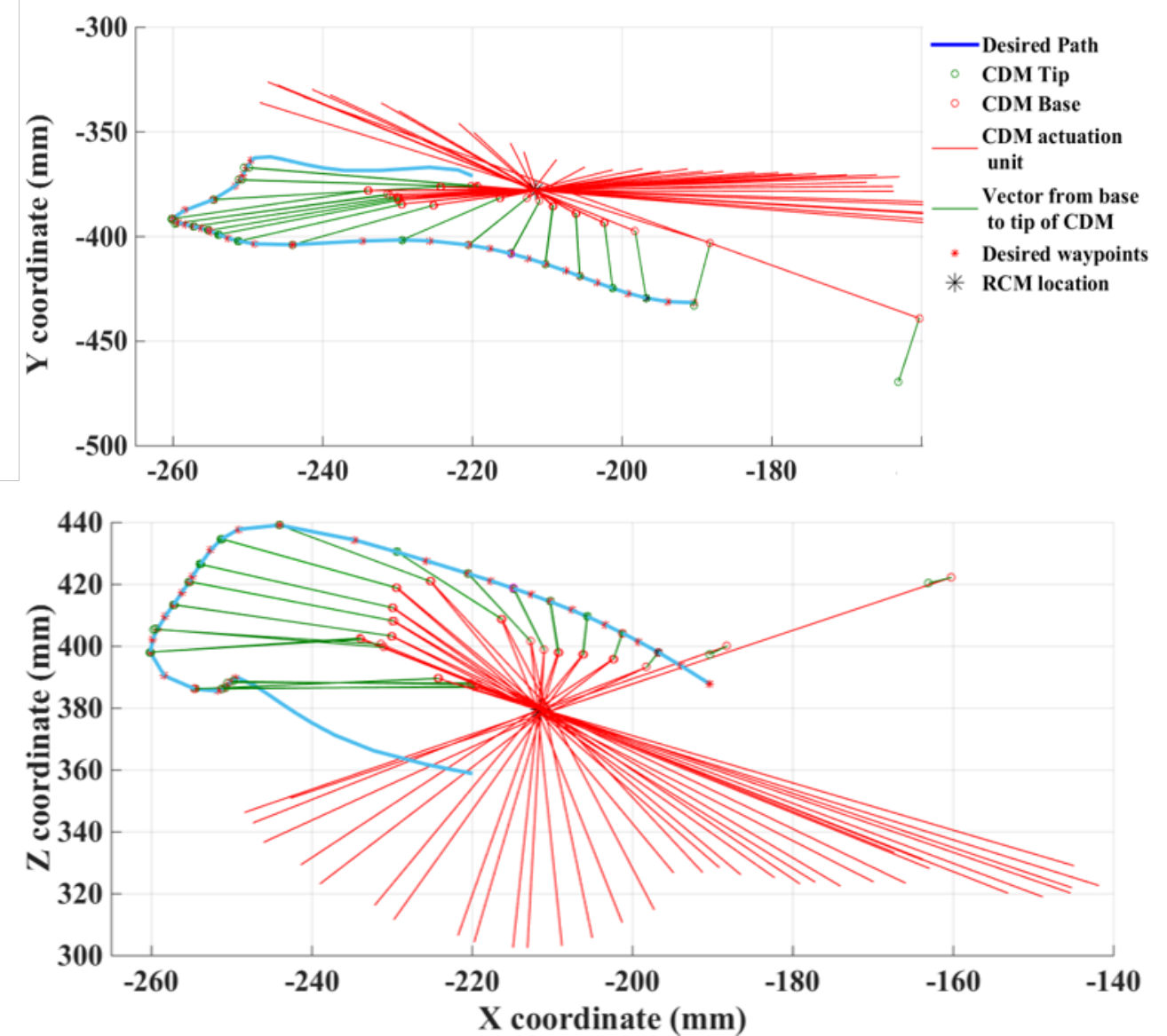}
	\caption{ADMM algorithm result for configuration and orientation of the integrated CDM and actuation unit in some points of the desired path.}
	\label{fig:ADMMa}
\end{figure}

Fig. \ref{fig:ADMMa} demonstrates the resulted configurations of the integrated system along the considered path when applying the ADMM algorithm. During this simulation the following parameters were used for evaluations: $\lambda=2e^{-4}$,   the maximum allowable tip error $(MTE=\epsilon_{CDM \ tip}\le0.5 mm)$ of the CDM defined as the maximum Euclidean distance between the desired position and the achieved CDM tip position,  and the maximum RCM constraint error  $(RCME=\epsilon_{RCM}\le0.5 mm)$, which defines the pivoting freedom of the CDM in the screw holes of the implant. Further, we used (\ref{eq:equation25}) to calculate optimal AL penalty parameter $\rho^*$ in each iteration.
%We compared the results of the ADMM with the active-set algorithm [28]- i.e. the \textit{lsqlin} function in MATLAB. This algorithm is used for solving constrained linear least-squares problems. We used the same simulation parameters for both algorithms and considered runtime and maximum number of tracked points on the assigned path as two criteria of evaluation. We ran each simulation three times and took the average as the runtime.
As shown, for each point, the integrated system could satisfy the constraints as well as the RCME and MTE criteria. The ADMM algorithm accomplished the task in 5.62 seconds ($\sim 0.15$ second/point). It should be noted that in [2], we solved the similar problem using an active-set algorithm with $RCME \leq 1 mm $ and achieved $\sim 4 mm$ error with a runtime of more than 300 seconds. The use of the ADMM, therefore, significantly improved the runtime and the rate of convergence.
%however, the \textit{lsqlin} function could only follow 2 way points in $>120$ seconds. Moreover, in [2], we investigated performance of the \textit{lsqlin} function in solving problem (\ref{eq:equation11}) without using the control algorithm \ref{control}- with $RCME\le1 mm$- and achieved about $4 mm$ error with a runtime more than 300 seconds. The use of the ADMM, therefore, significantly improved the runtime and the rate of convergence.
\subsection{Effect of Parameter $\rho$ on the Convergence Rate of the ADMM Algorithm}
\begin{figure}[!t]
	\centering
	\includegraphics[width=\linewidth]{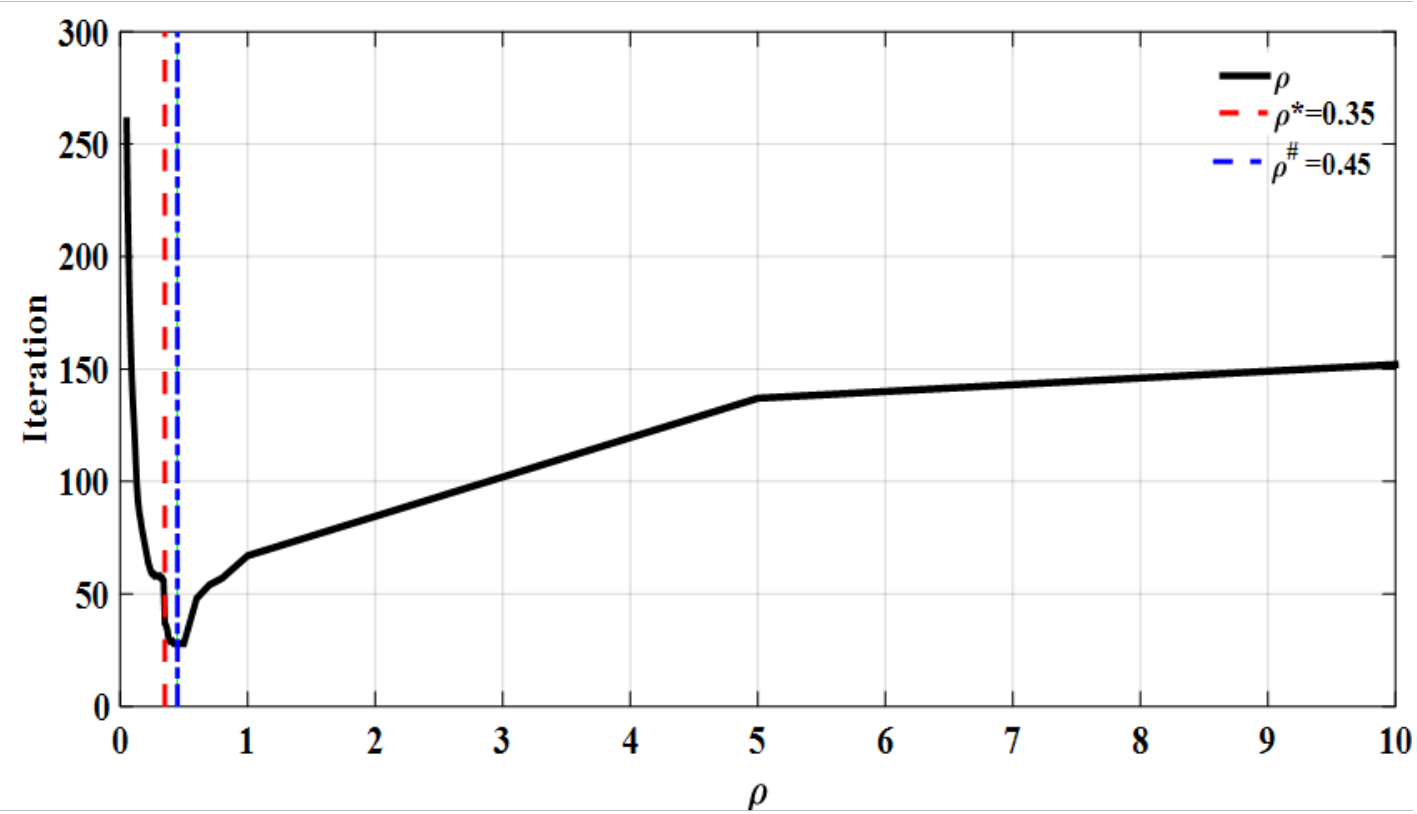}
	\caption{ Number of iterations required for the convergence of the ADMM algorithm considering 50 different penalty parameters$\rho \in [0.05,10]$  while keeping the other parameters constant. The dashed red line denotes the calculated optimal parameter$\rho^*$ and the dot-dashed blue line corresponds to the ideal parameter $\rho^\#$ obtained from the simulations.
	}
	\label{fig:rrho}
\end{figure}

\begin{figure}[!t]
	\centering
	\includegraphics[width=\linewidth]{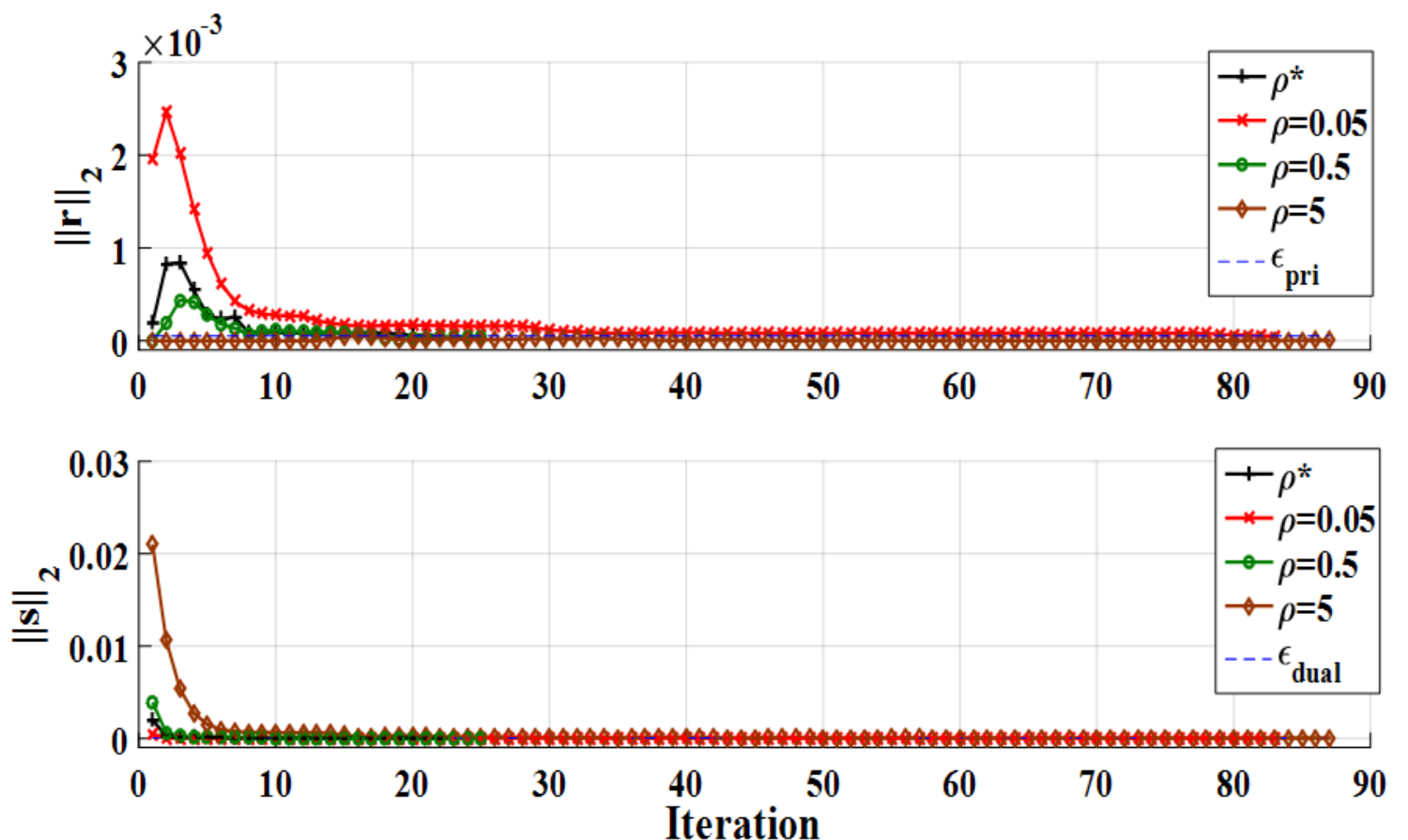}
	\caption{Convergence of the primal $r$ and dual $s$  residuals for four arbitrary considered AL penalty parameters $\rho$.
	}
	\label{fig:primals}
\end{figure}

\begin{figure}[!t]
	\centering
	\includegraphics[width=\linewidth]{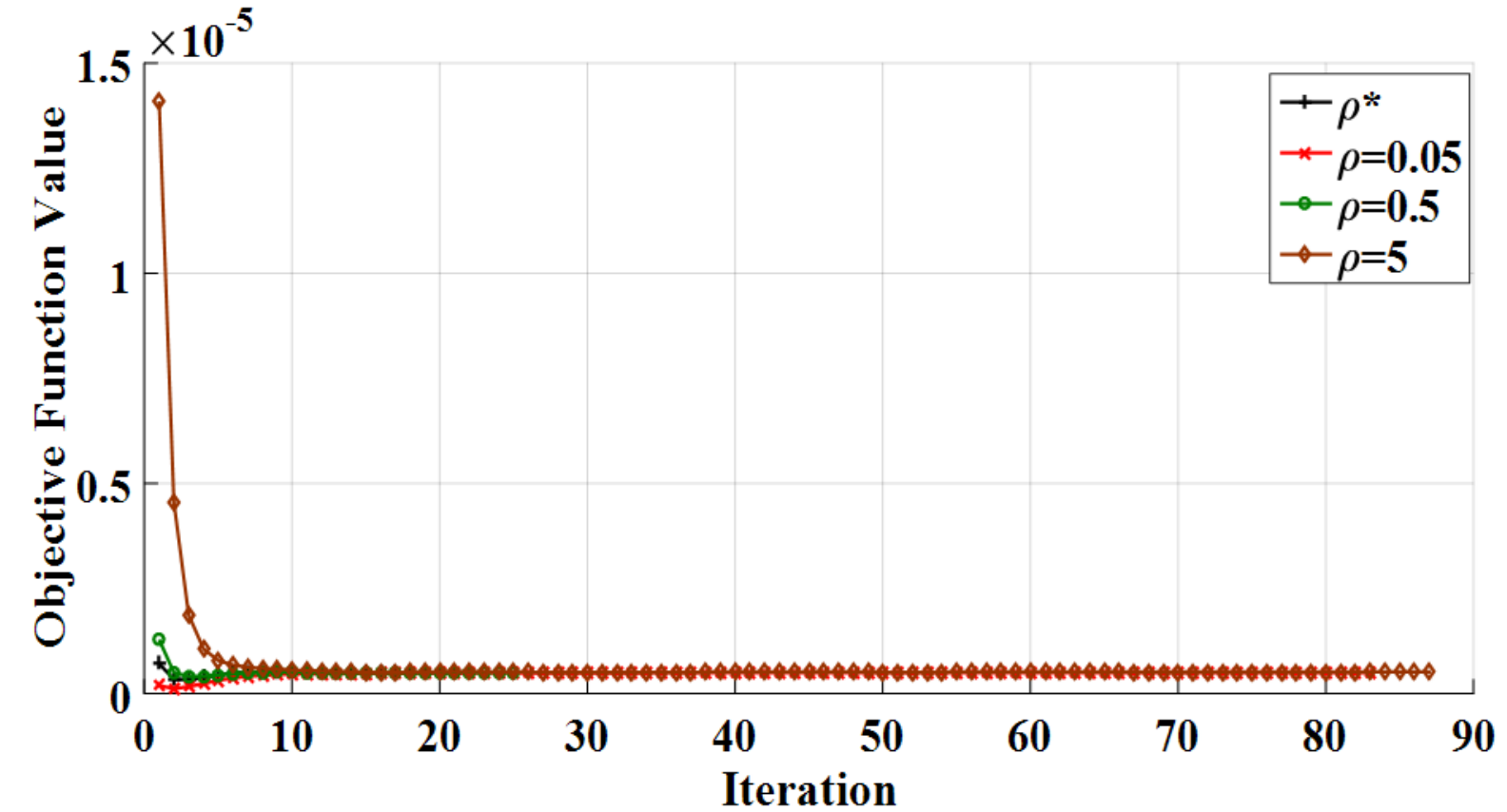}
	\caption{ Comparison of the objective function values for four arbitrary considered AL penalty parameters $\rho$.
	}
	\label{fig:objectives}
\end{figure}

As we proved in \textit{Theorem} 1, convergence of the ADMM algorithm under the mentioned assumptions is independent of parameter $\rho$. However, the convergence rate of the algorithm is directly affected by this parameter. To investigate this result, we ran the algorithm 50 times with different values of $\rho \in [0.05,10]$  while keeping the other parameters constant as: $\lambda=2e^{-4}, MTE\le 0.8 mm, \text{and}$  $RCME=1 mm$. We considered one target point and evaluated the convergence rate of the algorithm- defined as the maximum number of iterations- to reach this point from an initial configuration with a defined MTE. Furthermore, we used the optimal penalty parameter $\rho^*$ defined in (\ref{eq:equation25}) and evaluated the convergence rate of the algorithm. Fig. \ref{fig:rrho} presents the results of these experiments and shows the dependency of the convergence rate on the choice of the penalty parameter. As we proved in \textit{Theorem} 1, for all the 50 considered $\rho$, the algorithm converges; however, values larger than 0.6 and less than 0.3 significantly increase the convergence rate. Furthermore, using (\ref{eq:equation25}), we calculated the optimal penalty parameter  $\rho^*=0.35$ and compared it with the ideal value of   $\rho^{\#}=0.45$ obtained from the simulations (Fig.  \ref{fig:rrho}). As we discussed in Section \ref{convergence parameter}, for a tall matrix  A, $\rho^*$ does not return the optimal AL penalty parameter, however, it still provides a sufficiently close estimation of the optimal parameter and results in a fast convergence rate (37 iterations) compared to the ideal case  $\rho^\#$ (28 iterations). Fig. \ref{fig:primals} and Fig. \ref{fig:objectives} compare the primal and dual residuals and objective function values for four different cases of $\rho$, respectively. As we expected, for the case of  $\rho^*$, number of iterations is less than other considered penalty parameters.
\subsection{Sensitivity to the MTE and RCME}
The MTE defines the accuracy of the tracking as the maximum allowable Euclidean distance between the CDM tip and the desired point. On the other hand, the RCME is a measure assigning maximum allowable violation from the RCM constraint. Clearly, we are interested in the smallest values for these two parameters. As such, we considered various values for the RCME and the MTE and carried out simulations with a constant $\lambda=2e^{-4}$. Table \ref{tab:title} summarizes the results of these simulations. In all simulations, the ADMM successfully tracked the path and reached all 36 waypoints without violating considered MTEs. The maximum overall runtime using ADMM algorithm for tracking the considered trajectory consisting of 36 waypoints was 21.52 second ($\sim0.6$ second/point) and the minimum runtime was 4.27 second (~0.12 second/point). The average overall runtime for all 10 ADMM simulations in MATLAB, was 13.3 second and the runtime for each waypoint was about 0.37 second. Furthermore, the ADMM demonstrated acceptable sensitivity and robustness to the reduction of the tracking error (up to MTE= 0.1 mm) as well as the maximum violation of the RCM constraint (up to RCME= 0.5 mm).

\begin{table}
	\caption {RESULTS OF THE SIMULATIONS FOR THE CONSIDERED TRAJECTORY USING $\lambda = 2e^{-4}$} 
	\label{tab:title} 
	\begin{center}
		\begin{tabular}{c c c c}
			\hline
			MTE (mm) & RCME (mm) & run time (s) \\ [0.5ex] 
			\hline
			0.1 & 0.5 & 21.52 \\ 
			0.2 & 0.5 & 18.43 \\
			0.5 & 0.5 & 5.74 \\
			0.8 & 0.5 & 13.16 \\
			1 & 0.5 & 7.62 \\ [1ex] 
			\hline
			0.1 & 1 & 16.72 \\ 
			0.5 & 1 & 24.98 \\
			0.8 & 1 & 6.18 \\
			1 & 1 & 14.37 \\
			1.2 & 1 & 4.27 \\ [1ex] 
			\hline
		\end{tabular}
	\end{center}
\end{table}
% \addtolength{\textheight}{-2cm}   % This command serves to balance the column lengths

%%%%%%%%%%%%%%%%%%%%%%%%%%%%%%%%%%%%%%%%%%%%%%%%%%%%%%%%%%%%%%%%%%%%%%%%%%%%%%%%
\section{Concluding Remarks}
In this paper, we formulated the constrained redundancy resolution problem of manipulators in the context of a constrained $\ell_2$-regularized quadratic minimization problem. We solved this problem using a generic technique of convex optimization called ADMM algorithm. Further, we proved global convergence of the algorithm at linear rate and introduced expressions and assumptions for the involved parameters that results in a fast convergence rate. We validated the analytical results using simulation on a novel redundant surgical workstation. 
In conclusion, we demonstrated  the global convergence and robustness of the optimally tuned ADMM algorithm independent of the choice of parameters involved in the constrained DLM formulation. 
Future efforts will focus on validating the proposed workstation and method on other applications such as treatment of articular cartilage injury \cite{behrou2018numerical}, \cite{behrou2017novel} and osteonecrosis of femoral head \cite{alambeigi2017curved}. Further, we plan to compare the performance of the proposed optimally tuned ADMM with other redundancy resolution methods in the literature. 

\bibliographystyle{./bib/IEEEtran}
\bibliography{./bib/references}

\begin{thebibliography}{10}
\providecommand{\url}[1]{#1}
\csname url@rmstyle\endcsname
\providecommand{\newblock}{\relax}
\providecommand{\bibinfo}[2]{#2}
\providecommand\BIBentrySTDinterwordspacing{\spaceskip=0pt\relax}
\providecommand\BIBentryALTinterwordstretchfactor{4}
\providecommand\BIBentryALTinterwordspacing{\spaceskip=\fontdimen2\font plus
\BIBentryALTinterwordstretchfactor\fontdimen3\font minus
  \fontdimen4\font\relax}
\providecommand\BIBforeignlanguage[2]{{%
\expandafter\ifx\csname l@#1\endcsname\relax
\typeout{** WARNING: IEEEtran.bst: No hyphenation pattern has been}%
\typeout{** loaded for the language `#1'. Using the pattern for}%
\typeout{** the default language instead.}%
\else
\language=\csname l@#1\endcsname
\fi
#2}}

\bibitem{simaan2009design}
N.~Simaan, K.~Xu, W.~Wei, A.~Kapoor, P.~Kazanzides, R.~Taylor, and P.~Flint,
  ``Design and integration of a telerobotic system for minimally invasive
  surgery of the throat,'' \emph{The International journal of robotics
  research}, vol.~28, no.~9, pp. 1134--1153, 2009.

\bibitem{kapoor2005suturing}
A.~Kapoor, N.~Simaan, and R.~H. Taylor, ``Suturing in confined spaces:
  Constrained motion control of a hybrid 8-dof robot,'' in \emph{Advanced
  Robotics, 2005. ICAR'05. Proceedings., 12th International Conference
  on}.\hskip 1em plus 0.5em minus 0.4em\relax IEEE, 2005, pp. 452--459.

\bibitem{patel2015evaluation}
N.~Patel, C.~A. Seneci, J.~Shang, K.~Leibrandt, G.-Z. Yang, A.~Darzi, and
  J.~Teare, ``Evaluation of a novel flexible snake robot for endoluminal
  surgery,'' \emph{Surgical endoscopy}, vol.~29, no.~11, pp. 3349--3355, 2015.

\bibitem{coemert2016integration}
S.~Coemert, F.~Alambeigi, A.~Deguet, J.~Carey, M.~Armand, T.~Lueth, and
  R.~Taylor, ``Integration of a snake-like dexterous manipulator for head and
  neck surgery with the da vinci research kit,'' in \emph{Proc. Hamlyn Symp.
  Med. Robot.}, 2016, pp. 58--59.

\bibitem{alambeigi2016design}
F.~Alambeigi, S.~Sefati, R.~J. Murphy, I.~Iordachita, and M.~Armand, ``Design
  and characterization of a debriding tool in robot-assisted treatment of
  osteolysis,'' in \emph{Robotics and Automation (ICRA), 2016 IEEE
  International Conference on}.\hskip 1em plus 0.5em minus 0.4em\relax IEEE,
  2016, pp. 5664--5669.

\bibitem{alambeigi2016toward}
F.~Alambeigi, Y.~Wang, R.~J. Murphy, I.~Iordachita, and M.~Armand, ``Toward
  robot-assisted hard osteolytic lesion treatment using a continuum
  manipulator,'' in \emph{Engineering in Medicine and Biology Society (EMBC),
  2016 IEEE 38th Annual International Conference of the}.\hskip 1em plus 0.5em
  minus 0.4em\relax IEEE, 2016, pp. 5103--5106.

\bibitem{sefati2016fbg}
S.~Sefati, F.~Alambeigi, I.~Iordachita, M.~Armand, and R.~J. Murphy,
  ``Fbg-based large deflection shape sensing of a continuum manipulator:
  Manufacturing optimization,'' in \emph{SENSORS, 2016 IEEE}.\hskip 1em plus
  0.5em minus 0.4em\relax IEEE, 2016, pp. 1--3.

\bibitem{8234018}
S.~Sefati, M.~Pozin, F.~Alambeigi, I.~Iordachita, R.~H. Taylor, and M.~Armand,
  ``A highly sensitive fiber bragg grating shape sensor for continuum
  manipulators with large deflections,'' in \emph{2017 IEEE SENSORS}, Oct 2017,
  pp. 1--3.

\bibitem{alambeigi2014control}
F.~Alambeigi, R.~J. Murphy, E.~Basafa, R.~H. Taylor, and M.~Armand, ``Control
  of the coupled motion of a 6 dof robotic arm and a continuum manipulator for
  the treatment of pelvis osteolysis,'' in \emph{Engineering in Medicine and
  Biology Society (EMBC), 2014 36th Annual International Conference of the
  IEEE}.\hskip 1em plus 0.5em minus 0.4em\relax IEEE, 2014, pp. 6521--6525.

\bibitem{chirikjian1995kinematics}
G.~S. Chirikjian and J.~W. Burdick, ``The kinematics of hyper-redundant robot
  locomotion,'' \emph{IEEE transactions on robotics and automation}, vol.~11,
  no.~6, pp. 781--793, 1995.

\bibitem{sen2003variational}
S.~Sen, B.~Dasgupta, and A.~K. Mallik, ``Variational approach for
  singularity-free path-planning of parallel manipulators,'' \emph{Mechanism
  and Machine Theory}, vol.~38, no.~11, pp. 1165--1183, 2003.

\bibitem{nakamura1987optimal}
Y.~Nakamura and H.~Hanafusa, ``Optimal redundancy control of robot
  manipulators,'' \emph{The International Journal of Robotics Research},
  vol.~6, no.~1, pp. 32--42, 1987.

\bibitem{hollerbach1987redundancy}
J.~Hollerbach and K.~Suh, ``Redundancy resolution of manipulators through
  torque optimization,'' \emph{IEEE Journal on Robotics and Automation},
  vol.~3, no.~4, pp. 308--316, 1987.

\bibitem{funda1993optimal}
J.~Funda, R.~H. Taylor, K.~Gruben, and D.~LaRose, ``Optimal motion control for
  teleoperated surgical robots,'' in \emph{Optical Tools for Manufacturing and
  Advanced Automation}.\hskip 1em plus 0.5em minus 0.4em\relax International
  Society for Optics and Photonics, 1993, pp. 211--222.

\bibitem{wilkening2017development}
P.~Wilkening, F.~Alambeigi, R.~J. Murphy, R.~H. Taylor, and M.~Armand,
  ``Development and experimental evaluation of concurrent control of a robotic
  arm and continuum manipulator for osteolytic lesion treatment,'' \emph{IEEE
  Robotics and Automation Letters}, vol.~2, no.~3, pp. 1625--1631, 2017.

\bibitem{bajo2012integration}
A.~Bajo, R.~E. Goldman, L.~Wang, D.~Fowler, and N.~Simaan, ``Integration and
  preliminary evaluation of an insertable robotic effectors platform for single
  port access surgery,'' in \emph{Robotics and Automation (ICRA), 2012 IEEE
  International Conference on}.\hskip 1em plus 0.5em minus 0.4em\relax IEEE,
  2012, pp. 3381--3387.

\bibitem{boyd2011distributed}
S.~Boyd, N.~Parikh, E.~Chu, B.~Peleato, and J.~Eckstein, ``Distributed
  optimization and statistical learning via the alternating direction method of
  multipliers,'' \emph{Foundations and Trends{\textregistered} in Machine
  Learning}, vol.~3, no.~1, pp. 1--122, 2011.

\bibitem{bertsekas2015convex}
D.~P. Bertsekas and A.~Scientific, \emph{Convex optimization algorithms}.\hskip
  1em plus 0.5em minus 0.4em\relax Athena Scientific Belmont, 2015.

\bibitem{khadem2016ultrasound}
M.~Khadem, C.~Rossa, R.~S. Sloboda, N.~Usmani, and M.~Tavakoli,
  ``Ultrasound-guided model predictive control of needle steering in biological
  tissue,'' \emph{Journal of Medical Robotics Research}, vol.~1, no.~01, p.
  1640007, 2016.

\bibitem{ghadimi2015optimal}
E.~Ghadimi, A.~Teixeira, I.~Shames, and M.~Johansson, ``Optimal parameter
  selection for the alternating direction method of multipliers (admm):
  quadratic problems,'' \emph{IEEE Transactions on Automatic Control}, vol.~60,
  no.~3, pp. 644--658, 2015.

\bibitem{horn2012matrix}
R.~A. Horn and C.~R. Johnson, \emph{Matrix analysis}.\hskip 1em plus 0.5em
  minus 0.4em\relax Cambridge university press, 2012.

\bibitem{murray1994mathematical}
R.~M. Murray, Z.~Li, S.~S. Sastry, and S.~S. Sastry, \emph{A mathematical
  introduction to robotic manipulation}.\hskip 1em plus 0.5em minus 0.4em\relax
  CRC press, 1994.

\bibitem{behrou2018numerical}
R.~Behrou, H.~Foroughi, and F.~Haghpanah, ``Numerical study of temperature
  effects on the poro-viscoelastic behavior of articular cartilage,''
  \emph{Journal of the mechanical behavior of biomedical materials}, vol.~78,
  pp. 214--223, 2018.

\bibitem{behrou2017novel}
------, ``A novel study of temperature effects on the viscoelastic behavior of
  articular cartilage,'' \emph{arXiv preprint arXiv:1710.05467}, 2017.

\bibitem{alambeigi2017curved}
F.~Alambeigi, Y.~Wang, S.~Sefati, C.~Gao, R.~J. Murphy, I.~Iordachita, R.~H.
  Taylor, H.~Khanuja, and M.~Armand, ``A curved-drilling approach in core
  decompression of the femoral head osteonecrosis using a continuum
  manipulator,'' \emph{IEEE Robotics and Automation Letters}, vol.~2, no.~3,
  pp. 1480--1487, 2017.

\end{thebibliography}

\end{document}